\documentclass[runningheads]{llncs}
\usepackage[table]{xcolor}

\definecolor{bestgreen}{HTML}{B7E4C7}
\definecolor{midgreen}{HTML}{D8F3DC}
\definecolor{loworange}{HTML}{FFE8CC}
\definecolor{midorange}{HTML}{FFD6A5}
\definecolor{headergray}{HTML}{F2F2F2}
\usepackage[T1]{fontenc}
\usepackage{graphicx}
\begin{document}
\title{Foundation Models Adaptation for Multi-View Multi-modal Cardiac MRI Segmentation and Direct Ejection Fraction Estimation}
\titlerunning{Foundation Models Adaptation for Multi-View Multi-modal Cardiac MRI}
%
\author{Sina Amirrajab \inst{1,2,3} \and
Cian Scannell \inst{3} \and
Volker Vehof \inst{2} \and
Michael Bietenbeck \inst{2} \and
Ali Yilmaz \inst{2}}

\authorrunning{S. Amirrajab et al.}
%
\institute{The D-Lab, Department of Precision Medicine, GROW - Research Institute for Oncology and Reproduction, Maastricht University, Maastricht, the Netherlands. \and
Division of Cardiovascular Imaging, Department of Cardiology I, University Hospital Münster,  Münster, Germany \\
 \and
Department of Biomedical Engineering, Eindhoven University of Technology, Eindhoven, The Netherlands}
\maketitle              

\begin{abstract}
Foundation models have shown strong transferability in cardiac MRI (CMR), but their effectiveness for heterogeneous multi-view and multi-sequence CMR analysis remains unclear. In this work, we explore the effectiveness of fine-tuning and combining different CMR foundation models for the Universal Multi-Sequence, Multi-Center and Multi-View CMR Segmentation (CMR-Multi) Challenge. CineMA was fine-tuned for cine and late gadolinium enhancement (LGE) segmentation across short-axis and long-axis views. For direct left-ventricular ejection fraction (LVEF) estimation, we used two recent frozen CMR foundation models to extract embedding vectors that were then combined using attention-based multiple-instance learning for LVEF regression. In the challenge validation set, cine segmentation achieved Dice scores of 0.862, 0.883, and 0.902 for short-axis, two-chamber and four-chamber cine MRI, respectively. LGE segmentation achieved Dice scores between 0.621 and 0.846 across views. The direct LVEF regression model achieved an MAE of 4.96 percentage points and a Pearson correlation of 0.91. These results indicate that foundation models can be effectively adapted and combined for multi-view CMR analysis, while accurate LGE scar segmentation remains a challenging task.

\end{abstract}
{\let\thefootnote\relax\footnotetext{The code is available at \url{https://github.com/sinaamirrajab/cmr-multi-cinema}}}

\section{Introduction}

Cardiovascular magnetic resonance imaging (CMR) provides complementary information on cardiac anatomy, function, and myocardial tissue characteristics. Cine CMR enables the assessment of ventricular motion and ejection fraction (EF), while late gadolinium enhancement (LGE) imaging identifies myocardial fibrosis and scar. However, automated analysis across multiple views and sequences remains challenging because conventional deep learning models are typically trained separately for each task, modality, and dataset.

Foundation models offer transferable representations that can be adapted to downstream tasks with limited labels. Recent CMR foundation models have demonstrated promising performance in segmentation, disease classification, functional assessment, and image-text representation learning. These include self-supervised multi-sequence CMR pretraining~\cite{jacob2025cmrfm}, CineMA multi-view cine CMR foundation model ~\cite{fu2026cinema}, the ViTa visual-tabular framework ~\cite{zhang2025vita}, and vision-language CMR models such as the CMR Transformer~\cite{shad2026cmr} and CMRCLIP~\cite{nakashima2026cmrclip}.

Recent systems have also explored the integration of specialized CMR models for segmentation, functional quantification, tissue characterization, and cardiovascular disease evaluation~\cite{qu2026baaicardiacagent}. However, the effectiveness of adapting and combining publicly available foundation models for heterogeneous multi-sequence and multi-view CMR data remains insufficiently understood.

In this work, we explore the effectiveness of fine-tuning and combining CMR foundation models for the Universal Multi-Sequence, Multi-Center, and Multi-View CMR Segmentation Challenge~\cite{qu2026cmrmulti}. We fine-tune CineMA-based specialists for cine and LGE segmentation across short-axis and long-axis views. In parallel, we combine frozen CMR Transformer and CMRCLIP embeddings using attention-based multiple-instance learning for direct EF estimation. Our experiments assess whether complementary representations of the foundation-model can support integrated anatomical, scar, and functional CMR analysis.

\section{Method}
\label{sec:method}

We propose a task-specialist framework for multi-view and multi-modal cardiac MRI analysis in the CMR-Multi challenge. The framework processes cine MRI acquired in short-axis (SAX), two-chamber (2CH), and four-chamber (4CH) views, and late gadolinium enhancement (LGE) MRI acquired in SAX, 2CH, 4CH views. It produces anatomical segmentation masks for cine MRI, anatomical and scar segmentation masks for LGE MRI, including right atrium segmentation (RAS) ~\cite{zhu2024ras,bai2025benchmark}, and a direct estimate of left-ventricular ejection fraction (LVEF) using cine images.

\begin{figure}
    \centering
    \includegraphics[width=1\linewidth]{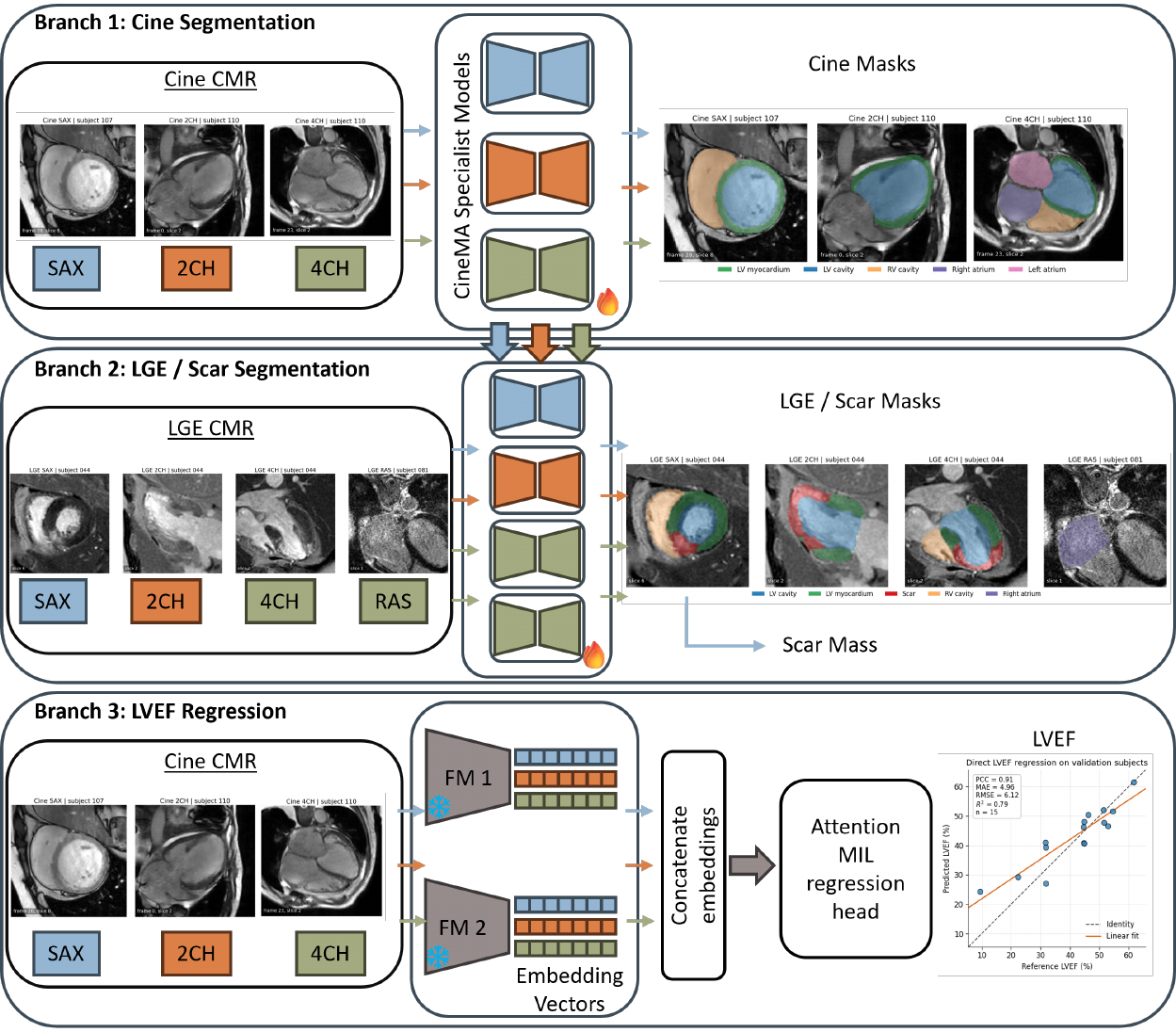}
    \caption{Proposed foundation model adaptation framework for multi-view, multi-modality CMR segmentation and EF prediction. Cine MRI views are processed by fine-tuned CineMA segmentation specialists, while LGE MRI views are processed by LGE/scar specialists initialized from the corresponding cine models. Frozen CMR Transformer and CMRCLIP encoders extract cine embeddings, which are concatenated and passed to an attention MIL regression head for LVEF prediction. The framework outputs cine masks, LGE/scar masks, scar mass, and LVEF.}
    \label{fig:placeholder}
\end{figure}

As shown in Figure \ref{fig:placeholder}, the method contains three complementary branches: cine segmentation, LGE/scar segmentation, and direct LVEF regression. Segmentation branches are trained as task specialists because target label sets, imaging planes, and image contrasts differ between cine and LGE acquisitions. The LVEF branch is independent of the segmentation masks and instead uses global cine representations from pretrained foundation models.

\subsection{Cine Segmentation}

Cine segmentation is performed with CineMA~\cite{fu2026cinema}, a pretrained CMR foundation model with released segmentation checkpoints. We train separate CineMA specialists for SAX, 2CH, and 4CH cine segmentation using the official patient-level training split. Each specialist is optimized for the label space of its corresponding view and predicts the relevant ventricular, myocardial, and atrial structures.

The SAX specialist is initialized from the released CineMA M\&Ms2 SAX segmentation checkpoint. The 4CH specialist is initialized from the released CineMA M\&Ms2 long-axis 4CH checkpoint. The 2CH specialist uses the same long-axis checkpoint as a view-compatible warm start, with the output head resized to the 2CH label space. When the released checkpoint label space differs from the challenge label space, anatomically matching output-head rows are copied, and unmatched rows are learned during supervised fine-tuning.

\subsection{LGE and Scar Segmentation}

The LGE branch transfers challenge-adapted cine specialists to LGE for scar segmentation. For each LGE task, we initialize a CineMA model from the corresponding fine-tuned cine model and resize the output head to the LGE label space. LGE SAX is initialized from the fine-tuned cine SAX specialist. LGE 2CH and 4CH are initialized by the corresponding fine-tuned cine long-axis specialists, and the model for RAS is initialized by the cine 4CH specialist because it provides the closest atrial representation.

This transfer strategy uses cine fine-tuning to adapt CineMA to the challenge anatomy before exposing the model to LGE contrast. Anatomically shared structures such as the LV cavity, myocardium, and RV cavity are transferred when available, while LGE-specific labels, including scar, are learned from LGE annotations. LGE models are therefore initialized with a spatially meaningful cardiac representation but remain fully supervised for the LGE task.

All segmentation specialists are trained with a combination of cross-entropy and soft Dice loss. All CineMA segmentation specialists are fine-tuned end-to-end with AdamW using a learning rate of \(10^{-4}\) for cine and \(5 \times 10^{-5}\) for LGE/scar models. Model selection is performed using validation loss with early stopping. Training, validation, and export use the official patient-level split.

\subsection{Direct LVEF Regression}

Direct estimation of LVEF is performed using two pretrained cine foundation models: CMR Transformer~\cite{shad2026cmr} and CMRCLIP~\cite{nakashima2026cmrclip}. The encoders are used as frozen feature extractors. For each subject, cine SAX, 2CH, and 4CH images are encoded into feature embeddings using both pretrained models. The embeddings are concatenated to form a combined representation for each cine instance.

Because each subject can contain a variable number of slices, frames, and views, we use an attention-based multiple-instance learning (MIL) regressor. The MIL attention head learns to aggregate the set of concatenated cine embeddings into a subject-level representation and predicts LVEF as a continuous value. Only the regression head is trained using training LVEF values. This direct EF branch is independent of the segmentation masks and can therefore exploit global appearance and motion information across the full cine study.

\subsection{Validation Results}

\begin{table}[t]
\centering

\caption{Validation segmentation comparison by modality and view. Green indicates the better method for each metric within each task. For DSC, higher is better; for HD95, lower is better.}
\label{tab:foundation-vs-scratch-macro-compact}
\resizebox{0.8\linewidth}{!}{%
\begin{tabular}{lrrrrr}

Task & N & \multicolumn{2}{c}{CineMA} & \multicolumn{2}{c}{Scratch nnU-Net} \\

 & & DSC (\%) $\uparrow$ & HD95 (mm) $\downarrow$ & DSC (\%) $\uparrow$ & HD95 (mm) $\downarrow$ \\

CINE SAX & 15
& \cellcolor{loworange} 86.2
& \cellcolor{loworange} 3.4
& \cellcolor{bestgreen}\textbf{89.6}
& \cellcolor{bestgreen}\textbf{2.8} \\

CINE 2CH & 15
& \cellcolor{bestgreen}\textbf{88.3}
& \cellcolor{bestgreen}\textbf{2.0}
& \cellcolor{loworange} 87.3
& \cellcolor{loworange} 2.3 \\

CINE 4CH & 15
& \cellcolor{bestgreen}\textbf{90.2}
& \cellcolor{bestgreen}\textbf{1.8}
& \cellcolor{loworange} 89.2
& \cellcolor{loworange} 5.3 \\

LGE SAX & 7
& \cellcolor{bestgreen}\textbf{65.5}
& \cellcolor{bestgreen}\textbf{15.5}
& \cellcolor{loworange} 64.2
& \cellcolor{loworange} 21.3 \\

LGE 2CH & 7
& \cellcolor{bestgreen}\textbf{62.1}
& \cellcolor{loworange} 10.3
& \cellcolor{loworange} 57.4
& \cellcolor{bestgreen}\textbf{9.3} \\

LGE 4CH & 7
& \cellcolor{bestgreen}\textbf{74.6}
& \cellcolor{bestgreen}\textbf{4.4}
& \cellcolor{loworange} 69.3
& \cellcolor{loworange} 9.3 \\

LGE RAS & 14
& \cellcolor{bestgreen}\textbf{84.6}
& \cellcolor{bestgreen}\textbf{7.7}
& \cellcolor{loworange} 80.0
& \cellcolor{loworange} 11.8 \\

\end{tabular}%
}
\end{table}

Table \ref{tab:foundation-vs-scratch-macro-compact} compares the fine-tuned CineMA models with nnU-Net ~\cite{isensee2021nnu} models trained from scratch on the validation set. CineMA cine specialists achieved validation Dice scores of 0.862, 0.883 and 0.902 for SAX, 2CH, and 4CH cine MRI, respectively. These results indicate that CineMA representations transfer effectively to challenge-specific cine anatomical segmentation tasks. However, scratch nnU-Net achieved the best CINE SAX performance, with a Dice score of 0.896 and HD95 of 2.8 mm. In contrast, CineMA performed better for the long-axis cine views, improving both Dice and HD95 in 2CH and 4CH. This suggests that foundation-model pretraining was most beneficial for views with fewer training examples or less volumetric context.

LGE segmentation was more challenging due to the distinct image contrast and the small and heterogeneous appearance of the scar. LGE specialists achieved validation macro Dice scores of 0.655 for SAX, 0.621 for 2CH, 0.746 for 4CH and 0.846 for RAS. The lower scores for LGE SAX and 2CH reflect the difficulty in modeling enhancement and scar extent compared to cine anatomy.
Compared with scratch nnU-Net, CineMA improved Dice in all LGE views and reduced HD95 in SAX, 4CH, and RAS. The largest gains were observed for LGE 4CH and RAS, where CineMA improved both overlap and boundary accuracy. This indicates that pretrained cine representations can still provide useful initialization for LGE segmentation, although the modality shift remains substantial.

\begin{table}[t]
\centering

\caption{Validation LVEF comparison for direct EF regression and segmentation-derived EF from CineMA and scratch nnU-Net masks. Green indicates the best-performing method for each metric. For MAE, RMSE, and absolute bias, lower is better; for $R^2$, correlation, and within-threshold accuracy, higher is better.}
\label{tab:foundation-vs-scratch-ef-compact}
\resizebox{0.73\linewidth}{!}{%
\begin{tabular}{lccc}

Metric 
& Direct EF 
& CineMA Seg EF 
& nnU-Net Seg EF \\

MAE $\downarrow$ 
& \cellcolor{midgreen} 4.965 
& \cellcolor{loworange} 9.487 
& \cellcolor{bestgreen}\textbf{3.971} \\

RMSE $\downarrow$ 
& \cellcolor{midgreen} 6.125 
& \cellcolor{loworange} 11.970 
& \cellcolor{bestgreen}\textbf{5.225} \\

$R^2$ $\uparrow$ 
& \cellcolor{midgreen} 0.788 
& \cellcolor{loworange} 0.189 
& \cellcolor{bestgreen}\textbf{0.846} \\

PCC $\uparrow$ 
& \cellcolor{midgreen} 0.911 
& \cellcolor{loworange} 0.616 
& \cellcolor{bestgreen}\textbf{0.934} \\

Spearman $\uparrow$ 
& \cellcolor{midgreen} 0.846 
& \cellcolor{loworange} 0.543 
& \cellcolor{bestgreen}\textbf{0.896} \\

\end{tabular}%
}
\end{table}

Table \ref{tab:foundation-vs-scratch-ef-compact} shows that segmentation-derived EF from scratch nnU-Net achieved the best overall LVEF accuracy, with an MAE of 3.97 EF percentage points. Direct EF regression was also competitive, reaching an MAE of 4.96 EF percentage points and showing strong correlation with the reference EF. In contrast, CineMA-derived EF showed weaker agreement, despite good segmentation performance in several views. The combined CMR Transformer and CMRCLIP attention MIL regressor achieved an RMSE of 6.12, ($R^2$) of 0.79, Pearson correlation of 0.91, and Spearman correlation of 0.85. Combining both foundation-model representations improved performance compared with the single-model baseline, which achieved a validation MAE of 5.27 EF percentage points.

Overall, these results support a task-specialist strategy rather than a single dominant model. CineMA was useful for long-axis cine and several LGE segmentation tasks, but scratch nnU-Net remained stronger for SAX cine segmentation and produced the most accurate segmentation-derived EF. This suggests that foundation-model initialization can be beneficial in selected settings, while conventional supervised segmentation remains highly competitive when sufficient task-specific labels are available. LGE scar segmentation remained the most difficult component, particularly in SAX and 2CH views, likely because scar occupies a small fraction of the image and shows variable enhancement patterns. Direct LVEF regression provided a competitive alternative to segmentation-derived EF and benefited from combining CMR Transformer and CMRCLIP representations, suggesting that the two pretrained encoders capture complementary cine information relevant to ventricular function.

\section{Conclusion}
\label{sec:conclusion}

We presented a task-specialist CMR-Multi framework that combines CineMA-based segmentation fine-tuning with direct foundation-model-based LVEF regression. The approach uses separate cine and LGE/scar specialists, transfers cine-adapted representations to LGE tasks, and estimates LVEF from combined frozen cine embeddings using attention MIL. This design provides a reproducible and modular solution for joint anatomical segmentation, scar assessment, and functional prediction in multi-view cardiac MRI. The validation results suggest that foundation-model representations can be useful in selected tasks, while nnU-Net remains a strong comparator, particularly for SAX cine segmentation and segmentation-derived EF.

%
%
%
%
\bibliography{bib}
\end{document}